%% file: acl_latex.tex
\documentclass[11pt]{article}

\usepackage[preprint]{acl}

\usepackage{times}
\usepackage{latexsym}

\usepackage[T1]{fontenc}

\usepackage[utf8]{inputenc}

\usepackage{microtype}

\usepackage{inconsolata}

\usepackage{graphicx}
\usepackage{amsmath,amssymb,booktabs}
\usepackage{placeins}

\title{Beyond K/V Caches: Cross-Model State Handoff in Hybrid Linear-Attention LLMs}

\author{
  Jiacheng Cai \quad
  Tao Zhang \quad
  Siqi Liang \quad
  Yun Dai \quad
  Ziqian Zeng\thanks{Corresponding author. Email:
  \texttt{zqzeng@scut.edu.cn}.} \\
  South China University of Technology, China
}

\ifdefined\pdfgentounicode
  \input{glyphtounicode}
\fi
\hypersetup{pdftitle={Beyond K/V Caches: Cross-Model State Handoff in Hybrid Linear-Attention LLMs}}

\begin{document}
\maketitle

\input{sections/0_abstract}
\input{sections/1_introduction}
\input{sections/2_background}

\input{sections/3_preliminaries}
\input{sections/4_method}
\input{sections/5_experiments}
\input{sections/6_conclusion}
\clearpage
\input{sections/7_limitations}

\bibliography{custom}

\clearpage
\appendix

\input{sections/8_appendix}

\end{document}

%% file: sections/0_abstract.tex
\begin{abstract}
Production deployments often use different models  to balance cost and quality through cascading, routing, and switching models during long conversations. Context accumulates across turns. Each switch requires the target to repeat prefill over this context. Existing methods transfer full-attention (FA) K/V caches across models. Hybrid linear-attention models also retain recurrent states (REC) and convolution histories (Conv). The interaction between Conv and REC complicates cross-model state transfer. Conv perturbations can change keys and values in recurrent layers and alter REC updates. Conv histories are overwritten, but changes in REC can still affect later outputs. More accurate REC reconstruction need not improve continuation. Building on this, we propose Hybrid State Transfer (HST) for all three state types. It uses structured REC maps and preserves history order during Conv alignment. We calibrate the adapters using the frozen target's outputs during continuation. FA cache map updates are merged offline.  Across five benchmarks, Qwen3.5 4B$\rightarrow$27B and 9B$\rightarrow$27B achieve mean task retentions of 89.88\% and 94.11\%, respectively. With source caches already available, separate timing on synthetic 16K inputs yields handoff body speedups of $21.66\times$ and $21.91\times$.
\end{abstract}


%% file: sections/1_introduction.tex






\section{Introduction}
\label{sec:introduction}

Production deployments often use models of different sizes within a family to balance cost and quality through cascading, routing, and switching models during long conversations \citep{heo2026crossmodel,ong2025routellm}.
The source model may already have processed the context.
In this case, target prefill repeats computation over the same tokens to build the target's inference state.
The cost of this repeated computation increases with context length.
Cross-model state transfer aims to reuse the source model's computation by converting the source state into a form usable by the target.
Hybrid linear-attention models also retain recurrent states (REC) and convolution histories (Conv).
The interaction between Conv and REC complicates cross-model state transfer.

Existing work on cross-model state transfer focuses on the K/V caches of full-attention (FA) models.
Affine mappings fitted in closed form convert these caches across models~\citep{heo2026crossmodel,qu2026cachebridge}.
Nonlinear correction modules are added to an affine K/V mapping and trained through a frozen target model using continuation cross-entropy and forward KL losses~\citep{qiu2026cachelink}.
The hybrid linear-attention models studied here also use REC to summarize history and Conv to store projected features within a finite window~\citep{yang2025gdn}.
Source and target models may differ in layer layout, recurrent head count, and convolution channel count.
Conv also affects key and value computation in recurrent layers, and these signals drive REC updates.
Errors in Conv conversion can therefore propagate into REC and affect subsequent target computation.

Our controlled interventions show that the effects of Conv perturbations can persist in REC and alter later outputs after the affected history slots have been overwritten.
REC reconstruction controls further show that lower local error need not improve full-model continuation.
These observations motivate \textbf{continuation compatibility}: the target must be able to use the transferred context and update its state effectively.
We propose Hybrid State Transfer (HST), a framework for transferring FA K/V caches, REC, and Conv between hybrid linear-attention models.
HST combines structural adaptation with functional calibration through the target model and keeps both models' pretrained weights fixed.
It uses pair-specific REC mappings with optional state-conditioned corrections and affine Conv alignment that preserves history order.
For FA, learned weight updates are merged offline into the base affine maps.
All states and position metadata are installed at a consistent boundary.

We evaluate the proposed HST framework using models from the Qwen3.5 family~\citep{qwen35modelcards} on five benchmarks covering commonsense reasoning, knowledge-based question answering, and mathematical reasoning.
For 4B$\rightarrow$27B and 9B$\rightarrow$27B, mean task retention reaches 89.88\% and 94.11\%, exceeding the strongest compared hybrid baseline by 3.56 and 7.10 percentage points, respectively.
Supplementary 4B$\leftrightarrow$9B experiments examine handoff with matching state layouts.
Separate timing measurements on synthetic 16K inputs with source caches already available show handoff body speedups of $21.66\times$ and $21.91\times$ for the two main directions, respectively.
These measurements exclude source prefill, the target model's output projection from hidden states to vocabulary scores, and answer generation.

Our main contributions are:
\begin{itemize}
\item \textbf{Conv and REC during continuation.}
In controlled interventions, the effects of Conv perturbations can persist in REC and alter later outputs beyond the convolution window.
Separate reconstruction diagnostics show lower REC reconstruction error alongside higher continuation KL.

\item \textbf{Hybrid State Transfer (HST).}
HST transfers FA K/V caches, REC, and Conv with position metadata at a common prefix boundary.
It uses pair-specific REC maps with optional state-conditioned corrections and preserves history order during affine Conv alignment.
REC mappings and FA cache maps are calibrated offline through a frozen target model, and learned updates to the FA cache maps are merged into the base affine maps before deployment.

\item \textbf{Task performance and handoff latency.}
Across five benchmarks, HST achieves higher mean task retention than the evaluated hybrid baselines for both 4B$\rightarrow$27B and 9B$\rightarrow$27B.
Separate timings on synthetic inputs with source caches already available show lower body latency for HST than for native target prefill.
\end{itemize}

%% file: sections/2_background.tex



\section{Related Work}
\label{sec:related-work}

\paragraph{Affine K/V mappings.}
Cross-Model selects supporting source layers and fits separate affine ridge maps for K and V. It removes source RoPE before K conversion and applies target RoPE afterward~\citep{heo2026crossmodel}. CacheBridge constrains source-head support and weights fitting by attention sensitivity while retaining affine deployment~\citep{qu2026cachebridge}. These approaches provide mappings for explicit full-attention caches. Additional persistent states require their own representations and correspondences.

\paragraph{Functional cache calibration.}
CacheLink adds nonlinear K/V corrections to a Ridge base and trains them through a frozen receiver using continuation cross-entropy and forward KL~\citep{qiu2026cachelink}. Receiver-side supervision and the distinction between reconstruction and downstream utility therefore have precedents. We examine continuation when recurrent memory and convolution histories also participate, and study full-attention cache calibration that can be folded into affine maps. The Ridge, CacheBridge, and CacheLink comparisons in Section~\ref{sec:experiments} use hybrid extensions implemented for this study.

%% file: sections/3_preliminaries.tex
\section{Preliminaries}
\label{sec:preliminaries}

\subsection{Hybrid Linear-Attention Models}
\label{sec:hybrid-models}

Linear attention admits recurrent computation with a fixed-size state~\citep{katharopoulos2020linear}.
Delta-rule formulations maintain a matrix memory and update it at each token~\citep{yang2024deltanet}.
The Qwen3.5 models studied here interleave full-attention (FA) layers with Gated DeltaNet (GDN) layers~\citep{qwen35modelcards,yang2025gdn}.

FA layers retain K/V caches indexed by token position, and the cache size grows with prefix length.
Each GDN layer retains a recurrent-state matrix (REC) and a convolution history (Conv).
REC summarizes previous updates, and Conv stores a finite window of projected input features in temporal order.
For a fixed model, REC and Conv sizes do not grow with prefix length.
Model pairs may differ in layer layout, recurrent head count, and convolution channel count.

At each recurrent step, a causal convolution combines the current projected features with Conv history to produce query, key, and value signals.
Separate input projections supply decay, write, and output gates.
The previous REC matrix and the current keys and values determine the REC update under the decay and write gates.
The query reads the updated matrix.
Normalization, output gating, and an output projection then produce the layer's hidden representation~\citep{yang2025gdn,qwen35modelcards}.
Conv therefore affects both the current readout and the information written into REC.
Effects written into REC can outlast Conv's finite window.

\subsection{Cross-Model State Transfer}
\label{sec:state-transfer}
\label{sec:background-state}

Let $C$ denote a prefix boundary after processing tokens $x_{1:C}$.
For model $m$, we represent the complete inference state at this boundary as
\begin{equation}
\mathcal{M}_m(C)
=
\left(
\{Z_{m,j}(C)\}_{j\in\mathcal{I}_m},
\pi_m(C)
\right).
\label{eq:prelim-state}
\end{equation}
Here, $\mathcal{I}_m$ indexes the persistent state components, and $Z_{m,j}(C)$ denotes one component.
The metadata $\pi_m(C)$ contains position indices, cache lengths, and attention masks.
All components and metadata describe the same processed prefix $x_{1:C}$.
After subsequent REC updates, truncating the FA cache cannot recover an earlier REC matrix.

We consider source and target models $s$ and $t$ from the same family, using the same input sequence and aligned prefix boundaries.
Both models' pretrained weights remain fixed.
A transfer function $F_\theta$ converts the source state into a target state:
\begin{equation}
\widehat{\mathcal{M}}_t(C)
=
F_\theta\!\left(\mathcal{M}_s(C)\right).
\label{eq:prelim-transfer}
\end{equation}
Here, $\theta$ denotes the transfer parameters.
The target processes the remaining task input from the installed state, followed by answer scoring or generation.

Offline fitting and diagnostics use native target states from prefill of the same prefix.
Online conversion uses only the source boundary state and saved transfer parameters, avoiding target prefix recomputation.

\subsection{Continuation Compatibility}
\label{sec:continuation-compatibility}

\begin{figure}[t]
  \centering
  \includegraphics[width=\columnwidth]{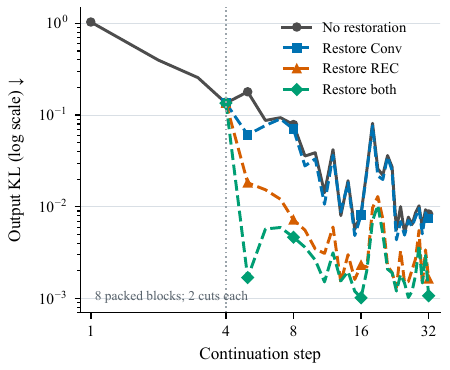}
  \caption{State restoration after zeroing Conv in native Qwen3.5-9B. Curves show all 32 steps, with two prefix boundaries averaged within each of eight packed blocks before averaging across blocks. Restore both denotes REC+Conv; FA caches are not restored. The dashed line marks restoration after measuring step~4. Both axes are logarithmic; KL is from native outputs to each branch.}
  \label{fig:conv-restoration}
\end{figure}

We use \emph{continuation compatibility} to describe the target model's ability to use transferred context and update its state effectively.
Task performance is the final criterion; preliminary diagnostics use forward KL from native outputs to diagnostic branches.
Figure~\ref{fig:conv-restoration} and Table~\ref{tab:rec-reconstruction} use different model configurations and inputs.

\paragraph{Conv perturbations.}
In native Qwen3.5-9B, we zero only Conv in eight packed token blocks, each with two prefix boundaries.
All branches consume the same 32 continuation tokens.
After measuring step~4, separate branches restore native REC, Conv, or both; FA caches are not restored.
At step~8, restoring REC yields output KL of $0.00721$, compared with $0.07023$ for restoring Conv.
The perturbed Conv slots have been overwritten, but effects can persist in REC and alter later outputs.
Appendix~\ref{app:conv-restoration} examines later fluctuations and an additional FA-restoration control.

\paragraph{REC reconstruction.}
A separate diagnostic uses an auxiliary 4B$\rightarrow$27B
configuration and 32 development documents.
These documents participated in candidate selection,
so this comparison is exploratory.
Base denotes the auxiliary transfer configuration before
the diagnostic reconstruction adjustments.
The two residual candidates modify REC while retaining
Base's FA caches and Conv histories.
Both candidates are fitted to reconstruct native target REC states.
For each evaluated REC layer, they use the same residual basis
and predict its coefficients by ridge regression.
\emph{\mbox{REC residual}} uses summaries of source REC states;
\emph{Context residual} additionally uses summaries of
source FA caches and Conv histories.
These reconstruction candidates are separate from HST's
functionally trained conditional REC correction.
Compared with Base, \mbox{REC residual} reduces the initial relative MSE
from $1.00671$ to $0.28380$, but increases full-model continuation KL
from $0.64829$ to $0.69142$.
Lower local REC reconstruction error therefore need not
produce lower continuation KL.
Table~\ref{tab:rec-reconstruction} reports the comparison.

\begin{table}[t]
  \centering
  \small
  \setlength{\tabcolsep}{3pt}
  \renewcommand{\arraystretch}{1.12}
  \begin{tabular*}{\columnwidth}{@{\extracolsep{\fill}}lrr@{}}
    \toprule
    Configuration &
    \begin{tabular}[c]{@{}r@{}}
      Initial REC\\relative MSE $\downarrow$
    \end{tabular} &
    \begin{tabular}[c]{@{}r@{}}
      Continuation\\KL $\downarrow$
    \end{tabular} \\
    \midrule
    Base             & 1.00671 & 0.64829 \\
    REC residual     & 0.28380 & 0.69142 \\
    Context residual & 0.24696 & 0.67841 \\
    \bottomrule
  \end{tabular*}
  \caption{
  Initial REC reconstruction error and full-model continuation KL
  for the auxiliary 4B$\rightarrow$27B configuration
  on 32 development documents used for candidate selection.
  Results are averaged over nine conditions per document:
  three prefix lengths (128, 512, and 2048 tokens)
  crossed with three continuation branches
  (\emph{natural}, \emph{revisit}, and \emph{topic shift}).
  All configurations consume the same fixed continuation tokens.
  Lower reconstruction error need not yield lower continuation KL.
}
  \label{tab:rec-reconstruction}
\end{table}

A separate state-replacement study uses 18 documents
in a fixed auxiliary 4B$\rightarrow$27B configuration
(Appendix~\ref{app:component-ablation}).
We replace selected components with native target states
for the same prefix and retain the other components
in their transferred form at the handoff boundary.
Composite KL equally weights the output KL at the end
of the prompt and its mean over the next eight positions.
Replacing REC or FA reduces composite KL from $0.59291$
to $0.10060$ or $0.33515$, respectively.
Replacing both yields $0.01664$, whereas replacing only
Conv yields $0.61845$.
These results measure replacement effects within the
transferred configuration and do not isolate independent
component contributions.
Together, the diagnostics motivate explicit adaptation
of REC and Conv and calibration through the target
model's subsequent outputs.
The next section presents HST.

%% file: sections/4_method.tex
\section{Method}
\label{sec:method}

HST combines complete-state mapping with functional calibration through a frozen target model. The mappings adapt FA K/V caches, REC matrices, and Conv histories to the target's layer, head, and channel layouts. Both models' pretrained weights remain fixed throughout.

\subsection{Handoff Boundary and State Consistency}
\label{sec:method-interface}

Let $x_{1:N}$ be the task prompt, and let $r$ denote the number of real prompt tokens reserved for target processing after installation. The handoff boundary is $C=N-r$. The source processes $x_{1:C}$ and exports its state at $C$. The target consumes the remaining prompt $x_{C+1:N}$ from the converted state. This suffix is part of the original task input.

\begin{figure*}[t]
  \centering
  \includegraphics[width=\textwidth]{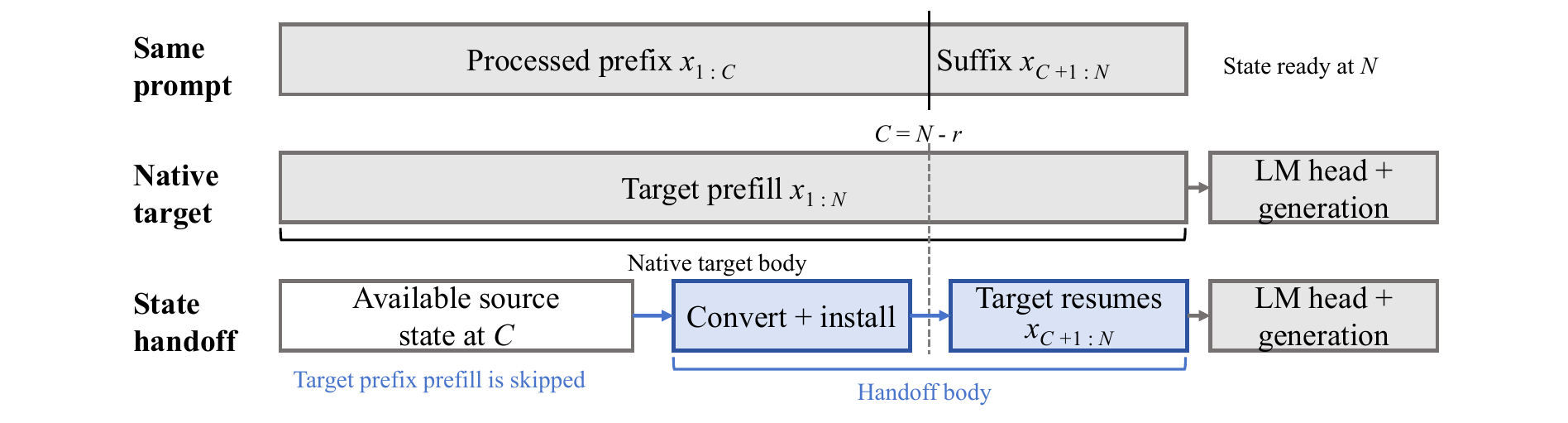}
  \caption{Handoff protocol and body latency definition. Given an available source state at $C=N-r$, handoff skips target prefill of $x_{1:C}$. The timed handoff includes state conversion, installation, the REC correction when used, and target suffix processing. The native reference processes the complete prompt. Both timed paths end when the target cache and final hidden state are ready. Source prefill, the vocabulary output projection, and answer generation are excluded. Block widths are schematic.}
  \label{fig:handoff-protocol}
\end{figure*}

Figure~\ref{fig:handoff-protocol} contrasts native prefill of $x_{1:N}$ with handoff from an available source state at $C$. Handoff skips target prefix prefill and resumes at $x_{C+1:N}$. Both paths reach $N$ before scoring or generation. Their final hidden states and next-token distributions can differ.

All components and position metadata must describe the same prefix $x_{1:C}$ (Section~\ref{sec:state-transfer}). We construct target FA, REC, and Conv states with consistent positions and cache lengths before the first target suffix token. Native target states provide offline fitting and supervision; online handoff uses the source state and trained adapters.

\subsection{Complete-State Mapping}
\label{sec:method-workflow}
\label{sec:method-base}

\begin{figure*}[t]
  \centering
  \includegraphics[width=\textwidth]{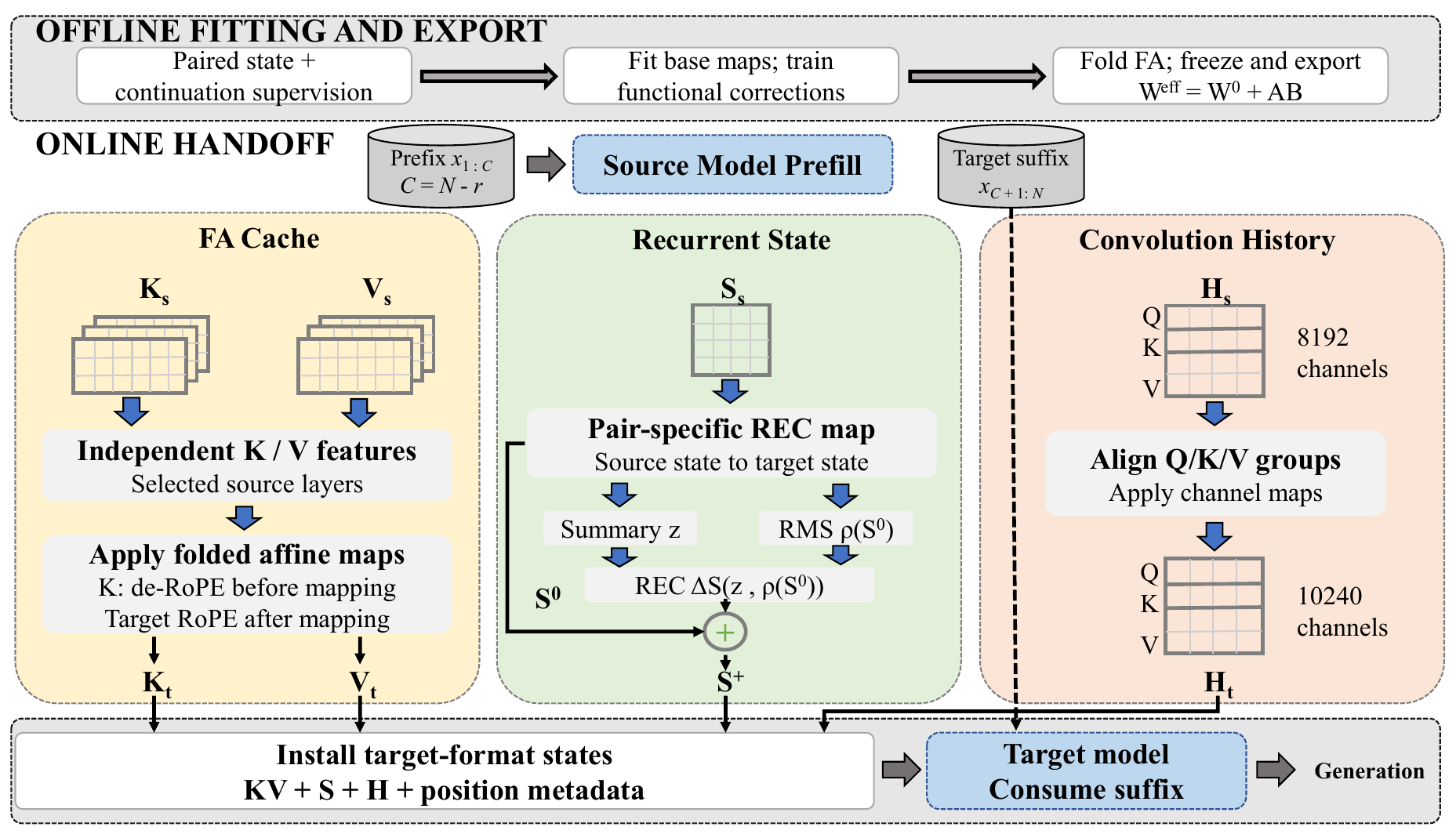}
  \caption{Complete-state handoff for Qwen3.5-4B$\rightarrow$27B with a state-conditioned REC correction. Offline training calibrates the REC mappings and trains the correction before the final FA-only stage. FA increments are merged into the cache-mapping weights before handoff. At $C=N-r$, the adapter converts FA K/V caches, REC matrices, and Conv histories. The REC correction produces $S^+$ from $S^0$ once. All target states and position metadata are ready before the target processes $x_{C+1:N}$.}
  \label{fig:state-handoff}
\end{figure*}

Source support layers are selected offline, and mapping parameters are fitted separately for each model pair. We omit the batch dimension and write feature vectors as row vectors. Figure~\ref{fig:state-handoff} illustrates the 4B$\rightarrow$27B configuration, including its state-conditioned REC correction. The 9B$\rightarrow$27B configuration uses its own calibrated mappings and does not apply this correction.

\paragraph{FA caches.}
For each target FA layer $j$, K and V share four selected source support layers but use separate feature vectors and affine maps. At historical position $i$, we first undo source RoPE on each source K head, then concatenate the resulting K features in the saved support-layer order to form $z^K_{j,i}$. We concatenate V features independently in the same order to form $z^V_{j,i}$. Within each layer, features follow the KV-head order and then the coordinate order within each head. The maps produce
\begin{equation}
\begin{aligned}
\widehat K_{j,i}&=\mathcal R_{t,i}\!\left(
\mathcal P_t(z^K_{j,i}W_j^K+b_j^K)\right),\\
\widehat V_{j,i}&=\mathcal P_t(z^V_{j,i}W_j^V+b_j^V),
\end{aligned}
\label{eq:fa-map}
\end{equation}
where $\mathcal P_t$ reshapes outputs into target KV heads and $\mathcal R_{t,i}$ applies target RoPE~\citep{heo2026crossmodel}. In the studied models, RoPE rotates the first 64 coordinates of each 256-dimensional K head. The remaining coordinates are unchanged.

Centered affine ridge regression initializes the K and V maps without regularizing their biases. The weights are shared across historical positions. The saved support-layer order is retained because the fitted coefficients correspond to that feature order. Mapping K after removing source RoPE separates the representation mapping from the explicit positional transformation.

\paragraph{REC matrices.}
Both model pairs use 16 groups defined by the native GDN Q/K heads. Each source group contains two value heads, and each target group contains three. For each target REC layer $j$, we select four source REC layers. At matrix coordinate $(a,b)$, the two heads from each selected source layer contribute eight values, concatenated into $\boldsymbol x_{j,g,a,b}$ for group $g$. The grouped affine map is
\begin{equation}
\boldsymbol s^0_{j,g,a,b}
=\boldsymbol x_{j,g,a,b}G_{j,g}+\beta_{j,g},
\label{eq:rec-group}
\end{equation}
where $G_{j,g}\in\mathbb R^{8\times3}$ and $\beta_{j,g}\in\mathbb R^3$. The output contains the values for the three target heads at the same matrix coordinate. Here, $\beta_{j,g}$ is the mapping bias, separate from the target model's recurrent gates.

The same weights and biases are applied at every coordinate of the $128\times128$ state matrices. This mapping combines selected source layers and heads without mixing matrix coordinates or native head groups. The two model pairs share this construction but use different support layers and parameters. Affine fitting initializes $(G,\beta)$, and offline functional calibration subsequently updates both. We assemble $S^0$ using these calibrated parameters; the superscript denotes the state before the conditional correction when used.

\paragraph{Conv histories.}
For each target recurrent layer, we use the first source layer in its saved REC support list. Conv stores the projected Q/K/V features before the recurrent layer's causal convolution. We divide these channels into Q, K, and V blocks and fit separate affine channel maps within their native groups. The same weights and biases are applied at all four history slots, without mixing slots. In the 4B$\rightarrow$27B configuration shown in Figure~\ref{fig:state-handoff}, this maps 8,192 source channels to 10,240 target channels.

The slots retain their original order from oldest to newest. Applying a shared channel map at each slot changes its feature representation while preserving its position in the convolution history. This alignment accounts for Conv's role in producing the keys and values used by subsequent REC updates.

\subsection{Functional Calibration and FA Folding}
\label{sec:method-calibration}

Low reconstruction error does not ensure that the target will use a transferred state effectively. We therefore calibrate the adapters through target continuation in separate offline stages: updating REC mappings, training the 4B$\rightarrow$27B REC correction, and learning an FA weight increment. The target model's pretrained parameters, including those used for recurrent state updates and the vocabulary output projection, remain fixed.

\paragraph{REC mapping calibration.}
Starting from the fitted affine maps, functional calibration updates both $G_{j,g}$ and $\beta_{j,g}$ through the frozen target. Initial calibration uses KL between native and transferred output distributions, together with normalized REC reconstruction error. Later task-oriented calibration also uses natural-text continuations and candidate-level outputs, including candidate-distribution KL and a mean-squared error term on centered candidate scores. Support layers and the grouped mapping structure remain fixed. The two model pairs are calibrated separately.

\paragraph{State-conditioned REC correction.}
The 4B$\rightarrow$27B configuration supplements $S^0$ with a correction conditioned on all 48 mapped target REC layers. At each layer, we compute row and column means, flatten each array in head order, and concatenate the two vectors. We concatenate the layer vectors in target-layer order and stop gradients through the resulting global summary. We partition it into 256-dimensional blocks, sum the blocks with alternating negative and positive signs, divide by the square root of the block count, and apply LayerNorm without trainable affine parameters. An affine projection followed by $\tanh$ produces coefficients $c_{j,h,k}$ for each target layer $j$, head $h$, and correction term $k$.

Let $\bar u_{j,h,k}$ and $\bar v_{j,h,k}$ be trainable direction vectors after separate L2 normalization. The corrected state is
\begin{equation}
S^+_{j,h}=S^0_{j,h}
+\gamma d\,\rho_{j,h}
\sum_{k=1}^{k_R}c_{j,h,k}
\bar u_{j,h,k}\bar v_{j,h,k}^{\top},
\label{eq:rec-patch}
\end{equation}
where $S^0_{j,h}\in\mathbb R^{d\times d}$, $d=128$, $\gamma=0.1$, and $k_R=2$. The scale $\rho_{j,h}$ is the RMS of the complete $d\times d$ matrix for that head, with gradients stopped through the scale computation. The added matrix has rank at most two; the base state $S^0$ has no imposed rank constraint.

The correction is trained separately with the REC maps, FA maps, Conv maps, and pretrained models fixed. At handoff, it is evaluated once before the target processes the prompt suffix. The target then continues with its native recurrent updates. The correction's coefficients and scale depend nonlinearly on the current mapped state, so it cannot be merged into a fixed affine REC map. The 9B$\rightarrow$27B configuration uses its calibrated $S^0$ directly.

\paragraph{Foldable FA calibration.}
Following the low-rank update parameterization of LoRA~\citep{hu2022lora}, the final calibration stage learns an increment to each K/V mapping weight:
\begin{equation}
W_j^{K/V}
=W_j^{0,K/V}+A_j^{K/V}B_j^{K/V}.
\label{eq:fa-lowrank}
\end{equation}
Both model pairs use rank-four increments. The base FA weights and biases remain fixed. Randomly initialized $A_j$ and zero-initialized $B_j$ give an initial increment of zero. REC mappings, the REC correction when used, and Conv mappings are also frozen during this stage. Only the increment is rank-constrained; the full base mapping matrices are retained.

The trained increment is independent of the transferred context. Before online handoff, it is merged into the base cache-mapping weight to obtain the $W_j^{K/V}$ used in Equation~\eqref{eq:fa-map}. The online FA branch therefore applies one affine map for K and one for V. This merge changes the cache mapper, while both models' pretrained weights remain fixed.

\subsection{Training and Online Handoff}
\label{sec:method-execution}

Base fitting uses paired source and native target states. REC mapping calibration and the separate 4B$\rightarrow$27B correction training precede final FA calibration.

Both task-oriented training of the conditional REC correction and final FA calibration use forward KL over 32 continuation prediction positions for natural-text supervision. Mathematical sequences use forward KL plus token cross-entropy with coefficient $0.25$ over 32 gold-solution prediction positions. Multiple-choice requests use forward KL over the complete candidate-score distribution plus correct-candidate cross-entropy with coefficient $0.25$. For the multiple-choice data in these stages, each candidate score is the sum of its token log-probabilities. A softmax over the scores gives the candidate distribution. Each multi-token candidate is evaluated from an independent prompt-state branch.

The native teacher is the frozen 27B target using its own prefix state. In the final FA stage of the 4B$\rightarrow$27B configuration, the first eight mathematical prediction positions instead use a frozen transfer teacher. This teacher uses the base FA maps, calibrated REC maps, aligned Conv histories, and the trained REC correction, without the final FA increment. The remaining 24 positions use the native target teacher. All 32 positions receive equal weight, and token cross-entropy uses the same gold solution throughout. This teacher schedule applies to mathematical sequences. The 9B$\rightarrow$27B configuration uses native target supervision at all 32 mathematical positions and has no conditional REC correction.

Online execution uses the saved parameters without optimization or teacher supervision:
\begin{enumerate}
  \item \textbf{Convert and install the boundary state.}
  Apply the saved FA, REC, and Conv maps to the source state at $C$ and assemble the target cache with consistent positions, cache lengths, and metadata.

  \item \textbf{Apply the configured REC correction.}
  In 4B$\rightarrow$27B, update the installed REC states from $S^0$ to $S^+$ once using the global summary. The 9B$\rightarrow$27B configuration retains $S^0$.

  \item \textbf{Resume target computation.}
  Process the real prompt suffix $x_{C+1:N}$ using the target's native update rules. At $N$, the target cache and final hidden state are ready. The language-modeling head then projects the hidden state into vocabulary scores for subsequent answer scoring or generation.
\end{enumerate}

The handoff body latency defined in Figure~\ref{fig:handoff-protocol} is evaluated in Section~\ref{sec:experiments-efficiency}.

FA conversion processes every retained K/V entry, so its cost grows with prefix length. REC and Conv have fixed sizes for a given model. State installation, the conditional correction when enabled, and target suffix computation also contribute to handoff cost. The latency benefit therefore depends on the target prefix computation saved and the cost of these remaining operations.


%% file: sections/5_experiments.tex
\section{Experiments}
\label{sec:experiments}


\subsection{Setup}
\label{sec:experiments-setup}

\paragraph{Models and hardware.}
We evaluate Qwen3.5-4B and Qwen3.5-9B as sources with Qwen3.5-27B as the target~\citep{qwen35modelcards}. We use two NVIDIA A800 GPUs with 80GB of memory each. Evaluation uses batch size 1, BF16 model weights and FA caches, and FP32 REC states. Both models' pretrained parameters remain fixed.

\paragraph{Tasks.}
We evaluate on ARC-Challenge (ARC-C)~\citep{clark2018arc}, HellaSwag~\citep{zellers2019hellaswag}, WinoGrande~\citep{sakaguchi2020winogrande}, MMLU~\citep{hendrycks2021mmlu}, and GSM8K~\citep{cobbe2021gsm8k}.
We use the full test splits of ARC-C (1,172 examples), MMLU (14,042 examples across 57 subjects), and GSM8K (1,319 examples), and the full validation splits of HellaSwag (10,042 examples) and WinoGrande (1,267 examples).
ARC-C, HellaSwag, and WinoGrande use 0-shot evaluation.
MMLU uses 5-shot prompts with demonstrations from each subject's development split, and GSM8K uses 8-shot chain-of-thought prompts.

\par\addvspace{6pt}
\noindent
\begin{minipage}{\linewidth}
\centering
\footnotesize
\setlength{\tabcolsep}{3pt}
\renewcommand{\arraystretch}{0.9}
\begin{tabular}{lrrrr}
\toprule
Task & Ridge & CacheBridge & CacheLink & Ours \\
\midrule
\multicolumn{5}{l}{4B$\rightarrow$27B} \\
ARC-C & 76.69 & 75.86 & 90.30 & \textbf{92.78} \\
HellaSwag & 84.55 & 83.16 & 92.38 & \textbf{94.64} \\
WinoGrande & 67.13 & 65.93 & 86.87 & \textbf{89.27} \\
MMLU & 66.05 & 61.88 & 74.73 & \textbf{81.05} \\
GSM8K & 49.41 & 5.74 & 87.33 & \textbf{91.66} \\
Avg. & 68.77 & 58.51 & 86.32 & \textbf{89.88} \\
\midrule
\multicolumn{5}{l}{9B$\rightarrow$27B} \\
ARC-C & 96.52 & 94.81 & 95.66 & \textbf{103.28} \\
HellaSwag & 91.21 & 91.21 & \textbf{99.33} & \textbf{99.33} \\
WinoGrande & 77.36 & 77.36 & \textbf{87.28} & \textbf{87.28} \\
MMLU & 83.66 & 84.16 & 67.93 & \textbf{88.22} \\
GSM8K & 77.29 & 23.79 & 84.86 & \textbf{92.42} \\
Avg. & 85.21 & 74.27 & 87.01 & \textbf{94.11} \\
\bottomrule
\end{tabular}
\captionof{table}{Task retention (\%, higher is better) relative to native 27B. Avg. equally weights task retentions. Bold marks the highest point estimates, including ties, per task and direction. Retention above 100\% indicates scores above native 27B. All methods use complete hybrid-state configurations.}
\label{tab:main-results}
\end{minipage}
\par\addvspace{6pt}

\par\addvspace{6pt}
\noindent
\begin{minipage}{\linewidth}
\centering
\makeatletter
\def\@captype{figure}
\makeatother
\includegraphics[width=0.88\linewidth]{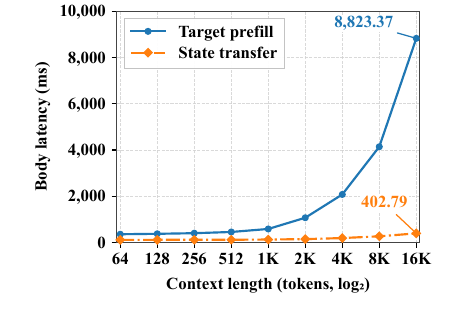}
\caption{Body latency comparison between state transfer and native target prefill for 9B$\rightarrow$27B on synthetic contexts with an available source cache. At 16K tokens, transfer reduces latency from 8,823.37 to 402.79 ms, yielding a $21.91\times$ speedup.}
\label{fig:handoff-body}
\end{minipage}
\par\addvspace{6pt}

\paragraph{Baselines and metrics.}
We compare complete hybrid-state extensions of Cross-Model~\citep{heo2026crossmodel}, CacheBridge~\citep{qu2026cachebridge}, and CacheLink~\citep{qiu2026cachelink}.
We refer to our Cross-Model extension as Ridge.
In 4B$\rightarrow$27B, Ridge and CacheBridge use closed-form REC maps, whereas CacheLink and our method use functionally calibrated REC adaptation.
All four methods use the same fitted Conv alignment maps for this direction.
For the 4B$\rightarrow$27B CacheLink extension, FA training uses native-target teachers with equally weighted token cross-entropy and forward KL; multiple-choice supervision uses token losses on gold candidate branches.
Task retention is $100\,s_{m,t}/s_{T,t}$, where $s_{m,t}$ and $s_{T,t}$ are method and native-target scores.
Avg. equally weights the five task retentions.
Absolute accuracies, ablations and state diagnostics, and matching-layout handoffs are reported in Appendices~\ref{app:absolute-scores}, \ref{app:ablations}, and \ref{app:compatible-directions}.

\subsection{Task Retention}
\label{sec:experiments-main}

Table~\ref{tab:main-results} shows mean retentions of 89.88\% and 94.11\% for our method in 4B$\rightarrow$27B and 9B$\rightarrow$27B, respectively, exceeding CacheLink by 3.56 and 7.10 percentage points. The gains are task-dependent: our method outperforms CacheLink on all five tasks in 4B$\rightarrow$27B, but ties on HellaSwag and WinoGrande in 9B$\rightarrow$27B. The latter direction's largest gain is on MMLU, where retention increases from 67.93\% to 88.22\%.

\subsection{Handoff Body Latency with an Available Cache}
\label{sec:experiments-efficiency}

We measure handoff body latency for 4B$\rightarrow$27B and 9B$\rightarrow$27B on synthetic contexts of 64--16K tokens with source caches already available. The timed transfer body includes state conversion, cache assembly and required copies, any REC correction, and target continuation until the target cache and hidden state are available. Timing excludes source prefill, the language modeling head, and answer generation. Speedup is computed as native target body latency divided by transfer body latency.

Figure~\ref{fig:handoff-body} compares native target prefill and state transfer times for 9B$\rightarrow$27B. Target prefill time grows more rapidly with context length than transfer time. At 16K, transfer reduces latency from 8,823.37 to 402.79 ms, yielding a $21.91\times$ speedup, compared with $3.01\times$ at 64 tokens.

Appendix~\ref{app:timing-details} reports timing records for both directions. For 4B$\rightarrow$27B, speedup increases from $2.78\times$ at 64 tokens to $21.66\times$ at 16K, where latency decreases from 8,644.23 to 399.02 ms.

%% file: sections/6_conclusion.tex
\section{Conclusion}
\label{sec:conclusion}

We propose a unified framework for cross-model state handoff in the Qwen3.5 hybrid model family to reduce repeated prefill when switching models. The framework combines structural adaptation of full-attention caches, recurrent states, and convolution histories with functional calibration through the target model, including foldable linear updates for attention-cache conversion. Across five benchmarks, 4B$\rightarrow$27B and 9B$\rightarrow$27B achieve mean task retentions of 89.88\% and 94.11\%, respectively. With source caches already available, the two directions achieve handoff body speedups of $21.66\times$ and $21.91\times$, respectively, on synthetic 16K contexts.

%% file: sections/7_limitations.tex
\section*{Limitations}
\label{sec:limitations}

We evaluate HST only within the Qwen3.5 family with Gated DeltaNet layers. Transfer to other linear-attention architectures, including
KDA~\citep{team2025kimi}, and to MoE models remains untested. Our benchmarks also provide limited coverage of long-context understanding. Task quality and handoff body latency are measured on benchmarks and synthetic contexts, respectively. Joint evaluation on the same inputs and under consistent inference settings is needed to establish the quality--efficiency trade-off.

%% file: sections/8_appendix.tex
\setcounter{dbltopnumber}{2}

\makeatletter
\setlength{\@dblfptop}{0pt}
\setlength{\@dblfpsep}{18pt}
\setlength{\@dblfpbot}{0pt plus 1fil}
\makeatother

\section{Supplementary Benchmark Results}
\label{app:absolute-scores}

Table~\ref{tab:absolute-scores} reports absolute accuracies for native models and all evaluated handoff configurations. Macro equally averages the five task accuracies.

\begin{table*}[tp]
\centering
\small
\setlength{\tabcolsep}{4pt}
\begin{tabular}{llrrrrrr}
\toprule
Model / direction & Method & ARC-C & HellaSwag & WinoGrande & MMLU & GSM8K & Macro \\
\midrule
4B & Native & 54.27 & 73.12 & 70.24 & 69.72 & 89.39 & 71.35 \\
9B & Native & 56.14 & 78.08 & 72.85 & 69.92 & 93.40 & 74.08 \\
27B & Native & 61.52 & 83.37 & 78.77 & 86.51 & 96.36 & 81.31 \\
\midrule
4B$\rightarrow$27B & Ridge & 47.18 & 70.49 & 52.88 & 57.14 & 47.61 & 55.06 \\
 & CacheBridge & 46.67 & 69.33 & 51.93 & 53.53 & 5.53 & 45.40 \\
 & CacheLink & 55.55 & 77.02 & 68.43 & 64.65 & 84.15 & 69.96 \\
 & Ours & 57.08 & 78.90 & 70.32 & 70.12 & 88.32 & \textbf{72.95} \\
\midrule
9B$\rightarrow$27B & Ridge & 59.38 & 76.04 & 60.94 & 72.37 & 74.48 & 68.64 \\
 & CacheBridge & 58.33 & 76.04 & 60.94 & 72.81 & 22.92 & 58.21 \\
 & CacheLink & 58.85 & 82.81 & 68.75 & 58.77 & 81.77 & 70.19 \\
 & Ours & 63.54 & 82.81 & 68.75 & 76.32 & 89.06 & \textbf{76.10} \\
\midrule
4B$\rightarrow$9B & Ridge & 56.83 & 75.35 & 68.59 & 71.91 & 84.53 & 71.44 \\
 & Ours & 55.97 & 76.09 & 71.51 & 69.72 & 88.32 & 72.32 \\
9B$\rightarrow$4B & Ours & 53.38 & 72.37 & 69.09 & 74.60 & 87.01 & 71.29 \\
\bottomrule
\end{tabular}
\caption{Absolute benchmark accuracy (\%). The native target provides the reference for each transfer direction. Macro equally averages tasks.}
\label{tab:absolute-scores}
\end{table*}

\subsection{Matching State Layouts}
\label{app:compatible-directions}

Table~\ref{tab:compatible-results} reports supplementary handoffs with matching state layouts and separately calibrated maps.

\begin{table*}[tp]
\centering
\small
\setlength{\tabcolsep}{4pt}
\begin{tabular}{lrrrrrr}
\toprule
Compatible configuration & ARC-C & HellaSwag & WinoGrande & MMLU & GSM8K & Avg. \\
\midrule
4B$\rightarrow$9B (Ridge) & 101.23 & 96.50 & 94.15 & 102.85 & 90.50 & 97.05 \\
4B$\rightarrow$9B (Ours) & 99.70 & 97.45 & 98.16 & 99.71 & 94.56 & 97.92 \\
\midrule
9B$\rightarrow$4B (Ours) & 98.36 & 98.97 & 98.36 & 107.00 & 97.34 & 100.01 \\
\bottomrule
\end{tabular}
\caption{Task retention (\%) for matching-layout handoffs. Native 9B and native 4B are the respective references for 4B$\rightarrow$9B and 9B$\rightarrow$4B. Avg. equally averages task retentions.}
\label{tab:compatible-results}
\end{table*}

\section{Ablation and Diagnostic Results}
\label{app:ablations}

These experiments use auxiliary 4B$\rightarrow$27B checkpoints, separate from the main comparison. The deployment ablation uses a native GSM8K reference of 1270/1319 (96.29\%); the main comparison uses 96.36\%.

\subsection{Deployment Ablation}
\label{app:absolute-ablation}
\label{sec:experiments-ablation}

Table~\ref{tab:deployment-ablation} toggles the FA increment and REC correction without retraining. The base FA, calibrated REC, and Conv maps remain fixed. Enabling both modules yields 88.32\% accuracy.

\begin{table*}[tp]
\centering
\small
\setlength{\tabcolsep}{4pt}
\begin{tabular}{lccrr}
\toprule
Configuration & FA increment & REC correction & Correct / total & Accuracy (\%) \\
\midrule
A & Off & Off & 832 / 1319 & 63.08 \\
B & Off & On & 878 / 1319 & 66.57 \\
C & On & Off & 1144 / 1319 & 86.73 \\
D & On & On & 1165 / 1319 & 88.32 \\
\midrule
Native 27B & --- & --- & 1270 / 1319 & 96.29 \\
\bottomrule
\end{tabular}
\caption{GSM8K deployment ablation in a fixed auxiliary configuration. Modules are toggled without retraining.}
\label{tab:deployment-ablation}
\end{table*}

\subsection{Native-State Replacement}
\label{app:component-ablation}

Table~\ref{tab:component-replacement} reports native-state replacement results on 18 documents. The selected components are replaced with native target states for the same prefix; all other components retain their transferred states at the handoff boundary. Composite KL equally weights the output KL at the end of the prompt and its mean over the next eight positions.

\begin{table*}[tp]
\centering
\small
\setlength{\tabcolsep}{4pt}
\begin{tabular}{lr}
\toprule
Native-state replacement & Composite KL $\downarrow$ \\
\midrule
None & 0.59291 \\
REC only & 0.10060 \\
FA only & 0.33515 \\
REC and FA & 0.01664 \\
Conv only & 0.61845 \\
\bottomrule
\end{tabular}
\caption{Native-state replacement on 18 documents. Only the specified components are replaced with native target states for the same prefix. Lower composite KL is better.}
\label{tab:component-replacement}
\end{table*}

\section{Handoff Latency and Speedup}
\label{app:timing-details}

Table~\ref{tab:handoff-timing} reports synthetic-input timings from 64 to 16K tokens. The native 27B reference is measured separately for each direction. With source caches already available, handoff body timing includes state conversion, installation and required copies, REC correction when used, and target suffix processing. Source prefill, the vocabulary output projection, and answer scoring or generation are excluded. These timing measurements are reported separately from benchmark quality results.

\begin{table*}[tp]
\centering
\small
\setlength{\tabcolsep}{5pt}
\begin{tabular}{rrrrrrr}
\toprule
 & \multicolumn{3}{c}{4B$\rightarrow$27B} & \multicolumn{3}{c}{9B$\rightarrow$27B} \\
\cmidrule(lr){2-4}\cmidrule(lr){5-7}
Input length & Native (ms) & Transfer (ms) & Speedup & Native (ms) & Transfer (ms) & Speedup \\
\midrule
64  & 367.43  & 132.20 & $2.78\times$  & 365.41  & 121.46 & $3.01\times$ \\
128 & 384.88  & 134.32 & $2.87\times$  & 383.93  & 123.83 & $3.10\times$ \\
256 & 410.29  & 135.71 & $3.02\times$  & 410.48  & 126.12 & $3.25\times$ \\
512 & 468.38  & 139.97 & $3.35\times$  & 462.58  & 127.43 & $3.63\times$ \\
1K  & 589.90  & 147.09 & $4.01\times$  & 593.08  & 136.37 & $4.35\times$ \\
2K  & 1068.88 & 162.24 & $6.59\times$  & 1077.29 & 154.58 & $6.97\times$ \\
4K  & 2068.30 & 198.62 & $10.41\times$ & 2082.30 & 200.31 & $10.40\times$ \\
8K  & 4114.88 & 269.78 & $15.25\times$ & 4143.11 & 275.72 & $15.03\times$ \\
16K & 8644.23 & 399.02 & $21.66\times$ & 8823.37 & 402.79 & $21.91\times$ \\
\bottomrule
\end{tabular}
\caption{Native target and handoff body times in milliseconds. K denotes 1024 tokens; speedup is native time divided by transfer time. Values are rounded to two decimal places.}
\label{tab:handoff-timing}
\end{table*}

\section{Training and Parameter Details}
\label{app:implementation}

\subsection{Training Configuration}

Experiments use two NVIDIA A800 GPUs with 80GB memory each. Evaluation uses batch size 1, BF16 model weights and FA caches, and FP32 REC states. Both pretrained backbones remain fixed.

The reported 4B$\rightarrow$27B base-fitting recipe uses 500 FineWeb-Edu~\citep{penedo2024fineweb} sequences of 1024 tokens, with four ordered source supports per FA and REC target layer. Conv maps share parameters across the four history slots. Both the weights and biases of the REC maps are functionally calibrated offline. The 4B$\rightarrow$27B REC correction has rank at most 2 and uses a 256-dimensional state summary; 9B$\rightarrow$27B uses its calibrated REC map directly. Both directions use rank-4 FA increments merged into the cache-mapping weights before online handoff.

During the final FA stage, the REC and Conv maps and the conditional REC correction, when used, remain fixed. The 4B$\rightarrow$27B configuration uses frozen transfer-teacher supervision for the first eight gold-solution prediction positions of mathematical sequences and native-target supervision for the remaining 24. The 9B$\rightarrow$27B configuration uses native-target supervision at all 32 mathematical prediction positions. Each final FA stage uses 384 training units and 768 updates. We use a learning rate of $3\times10^{-4}$ and evaluate the final checkpoint.

\subsection{Final FA-Stage Parameters and Training Time}
\label{app:parameter-details}
\label{sec:experiments-param}

Table~\ref{tab:training-cost} reports the additional FA parameters, update counts, and elapsed times for the two main transfer directions. Other state adapters remain fixed during this stage. Times exclude teacher preparation, earlier state calibration, and evaluation.

\begin{table*}[tp]
\centering
\small
\setlength{\tabcolsep}{6pt}
\begin{tabular}{lrrr}
\toprule
Configuration & Additional FA parameters (M) & Updates & Stage time (h) \\
\midrule
Main 4B$\rightarrow$27B & 0.655 & 768 & 2.92 \\
Main 9B$\rightarrow$27B & 0.655 & 768 & 3.36 \\
\bottomrule
\end{tabular}
\caption{Final FA-stage parameters, updates, and elapsed times for the main systems. Each stage uses 384 training units. Times cover final FA updates only.}
\label{tab:training-cost}
\end{table*}
\section{Native Conv Restoration Diagnostic}
\label{app:conv-restoration}

Native Qwen3.5-9B is evaluated on eight packed 1,024-token blocks from eleven source documents, at cuts 128 and 512. We average cuts within each block, then average blocks equally. Conv is zeroed at each cut. Each branch consumes the case's 32 true continuation tokens. After measuring step 4, selected REC and/or Conv states are restored once from the native reference at that step; FA is not restored. A further control restores REC, Conv, and FA together. The script verifies bitwise-equal native logits and zero KL for all $16\times28=448$ measurements at steps 5--32. Thus, FA caches retaining earlier perturbations can affect subsequent computation after REC and Conv have been restored. These oracle interventions are separate from transfer benchmarks.

\begin{samepage}
The REC+Conv branch reaches its highest post-restoration mean KL at step 19 ($0.01039322$; Table~\ref{tab:native-conv-restoration}). Block 2 (zero-based), cut 128, has KL $0.13518570$, contributing 81.29\% of this mean. The mean over the remaining 15 cases is $0.00207372$. The aggregate peak is dominated by this case, whose continuation remains within one source document.
\par
\end{samepage}

\begin{table}[t]
\centering
\small
\setlength{\tabcolsep}{4pt}
\begin{tabular}{rrr}
\toprule
Step & REC+Conv & REC+Conv+FA \\
\midrule
18 & 0.00703398 & 0.00000000 \\
19 & 0.01039322 & 0.00000000 \\
20 & 0.00508434 & 0.00000000 \\
\bottomrule
\end{tabular}
\caption{Output KL after oracle restoration in native Qwen3.5-9B. Both columns use the same eight blocks and two cuts per block.}
\label{tab:native-conv-restoration}
\end{table}